\documentclass[11pt]{article}

\usepackage{2_arXiv_Version/my_packages}
\usepackage{2_arXiv_Version/my_commands}
\usepackage{geometry}

\usepackage[numbers,sort&compress]{natbib}
\usepackage{xcolor}
\usepackage[
  colorlinks=true,
  linkcolor=blue,
  citecolor=teal,
  urlcolor=magenta
]{hyperref}

\graphicspath{{0_figures/}}
\title{\texttt{RTD-RAX}: Fast, Safe Trajectory Planning for Systems under Unknown Disturbances}

\author{
Evanns Morales-Cuadrado,
Long Kiu Chung,
Shreyas Kousik,
and Samuel Coogan%
\thanks{This work was supported in part by the National Science Foundation under awards \#2333488 and \#2219755 and in part by the Air Force Office of Scientific Research under Grant FA9550-23-1-0303. L.K. Chung was supported by NSF Award \#2449160.}%
\thanks{The authors are with the Georgia Institute of Technology, Atlanta, GA 30332, USA. E. Morales-Cuadrado and S. Coogan are with the School of Electrical and Computer Engineering; L. Chung and S. Kousik are with the School of Mechanical Engineering. S. Coogan is also with the School of Civil and Environmental Engineering.}%
\thanks{Emails: \texttt{\{egm, lchung33, shreyas.kousik@me, sam.coogan\}@gatech.edu}}
}

\date{}
\begin{document}
\maketitle

\begin{abstract}
Reachability-based Trajectory Design (RTD) is a provably safe, real-time trajectory planning framework that combines offline reachable-set computation with online trajectory optimization. However, standard RTD implementations suffer from two key limitations: conservatism induced by worst-case reachable-set overapproximations, and an inability to account for real-time disturbances during execution.
This paper presents \texttt{RTD-RAX}, a runtime-assurance extension of RTD that utilizes a non-conservative RTD formulation to rapidly generate goal-directed candidate trajectories, and utilizes mixed monotone reachability for fast, disturbance-aware online safety certification. When proposed trajectories fail safety certification under real-time uncertainty, a repair procedure finds nearby safe trajectories that preserve progress toward the goal while guaranteeing safety under real-time disturbances.
\end{abstract}


\section{Introduction}\label{sec:intro}
Safe motion planning for autonomous robots requires generating dynamically feasible trajectories in the presence of obstacles. In real-time receding-horizon settings and real-world deployment scenarios, planning must be fast enough to ensure persistent feasibility, account for \textit{a priori} unknown environmental disturbances, and guarantee that safe fallback maneuvers can be executed when a new safe trajectory cannot be found.
One framework that aims to address this challenge is Reachability-based Trajectory Design (RTD) --- a provably safe real-time motion planning framework that combines offline reachable-set computation with fast online trajectory optimization~\citep{SK-SV-FB-MJR-RV:2020} to achieve this task.

In RTD, a Forward Reachable Set (FRS) captures the behavior of a system tracking input-parameterized trajectories over a time horizon. At runtime, an online planning procedure senses obstacles, maps them onto the parameter space, and optimizes over the remaining set of safe trajectory parameters to achieve a desired objective. Moreover, all admissible trajectories are constructed such that if a new safe trajectory cannot be found quickly enough, or at all, the system can safely execute a pre-certified fallback maneuver, such as braking for ground vehicles or hovering for quadrotors~\citep{SK-SV-FB-MJR-RV:2020,SK-PH-RV:2019}.
The RTD framework has also been successfully applied to quadrotors~\citep{SK-PH-RV:2019}, robotic manipulators~\citep{ARMTD:PH-SK-BZ-DR-CB-MJR-RV:2020}, and adapted to train safe reinforcement learning agents~\citep{RTS:SSY-CC-SK-RV:2021}.

Despite its strengths, the RTD architecture has important limitations. In particular, parameterized reachable-set computation methods are too computationally expensive to be used
with high-fidelity, high-dimensional models~\citep{SK-SV-FB-MJR-RV:2020}. As a result, RTD relies on offline reachable-set computation using simplified, lower-dimensional models over sets of initial conditions that are later queried at runtime.

\begin{figure}[t]
    \centering
    \includegraphics[
        width=.65\textwidth,
        trim={.09cm .14cm 0.00cm .09cm},
        clip
    ]{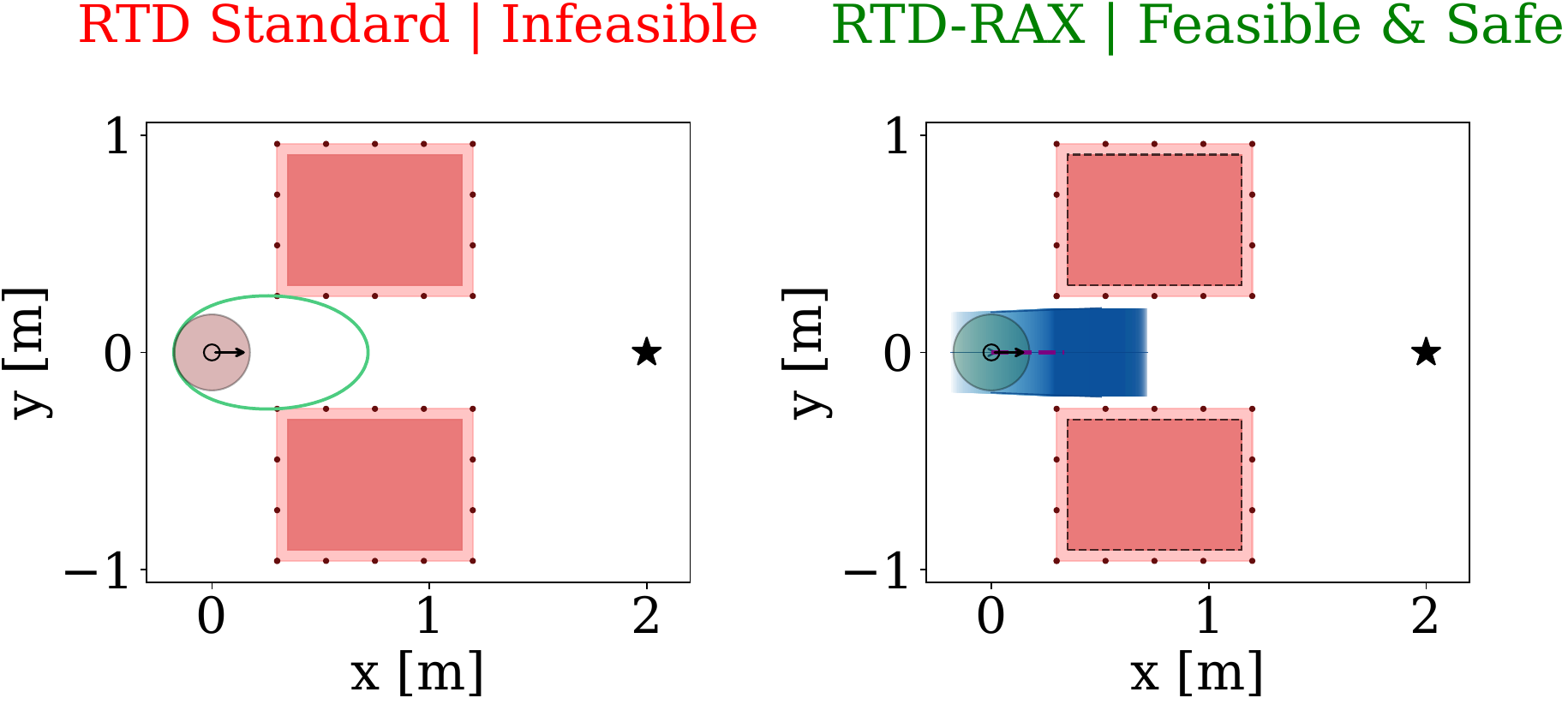}
\caption{
    Narrow-gap scenario; the robot is depicted as a circle in the starting position, goal as a star, obstacles as rectangles, the offline reachable set in green, and the online reachable set in blue.
    (a) Standard RTD: incorrectly classifies as infeasible.
    (b) \texttt{RTD-RAX}: feasible and certified safe by the mixed-monotone verifier.}
    \label{fig:gap_comparison}
\end{figure}

To account for the mismatch between the simplified planning model and the true system dynamics, the FRS is inflated using a worst-case bound on tracking error, typically estimated via sampling~\citep{SK-SV-FB-MJR-RV:2020}. The RTD safety certificate therefore inherits the conservatism of this bound.

This worst-case treatment of tracking error leads to conservative trajectory planning~\citep{SK-SV-RV:2017}. Although it provides provable safety in scenarios for which the RTD pipeline has previously been validated,
(\emph{e.g.}, in the absence of real-time disturbances),
it can cause trajectories that are in fact safe—and potentially optimal with respect to the planning objective—to be incorrectly classified as unsafe, resulting in unnecessarily cautious behavior such as rejecting efficient trajectories or triggering avoidable fail-safe maneuvers.
Furthermore, because the reachable sets are computed offline, the framework must rely on a lower-level controller to handle \textit{a priori} unknown environmental disturbances \citep{ARMOUR:JM-PH-BZ-CC-BW-SS-TZ-SD-SK-RV:2025}, which further increases conservativeness.

These observations motivate the use of a separate mechanism for execution-time safety verification in the RTD pipeline. By delegating safety certification to an online verification layer, the FRS used by RTD can be constructed with reduced conservatism, allowing the planner to consider trajectories that would otherwise be rejected under worst-case tracking-error inflation. Moreover, a fast-enough online verification layer may also be able to measure and account for real-time uncertainty in the system dynamics to ensure safety under bounded disturbances.

As such, this paper proposes~\texttt{RTD-RAX}, a \texttt{R}untime-\texttt{A}ssurance e\texttt{X}tension of RTD that employs mixed-monotone reachability (MMR) for online safety certification and re-planning. The key idea is a complementary architecture in which RTD's offline parameterized reachable sets are used solely for rapid candidate generation, without conservative tracking-error inflation, while safety is certified online using MMR for the specific closed-loop trajectory under current disturbance and uncertainty measurements. This enables, for the first time, safety certification for an RTD-based framework under \textit{a priori} unknown disturbances at runtime. Lastly, when a candidate cannot be certified as safe, a repair procedure modifies it to obtain a certifiably safe alternative.

We validate this architecture through three experiments: a narrow-gap scenario that the standard RTD framework would classify as unsafe; an angled-obstacle scenario in which an unsafe trajectory is caught before execution and re-planned; and a scenario demonstrating robustness to \textit{a priori} unknown disturbances that the standard RTD framework does not accommodate.

The remainder of the paper is organized as follows. Section~\ref{sec:literature_review} reviews the relevant RTD and mixed-monotone reachability literature, while Section~\ref{sec:RTD-RAX} describes their proposed integration into a cohesive safe planning framework. Section~\ref{sec:experiments} presents experimental results and comparisons with the traditional RTD framework, and Section~\ref{sec:conclusion} concludes the paper.
\footnote{Our code, documented results, and video demonstrations can also be found at: \url{https://evannsm.github.io/ws_RTD}}

\section{Literature Review}\label{sec:literature_review}
We review details from the relevant literature in Reachability-based Trajectory Design (RTD), and Mixed Monotone Reachability (MMR).

\subsection{Reachability-Based Trajectory Design}
\label{subsec:LR-RTD}

A representative RTD construction begins with a high-fidelity closed-loop system
\begin{equation}
    \label{eq:high_fid}
    \dot{x}_{\mathrm{hi}}(t) = f_{\mathrm{hi}}(t,x_{\mathrm{hi}}(t),u(t,x_{\mathrm{hi}}(t))),
\end{equation}
where \(x_{\mathrm{hi}}(t) \in X_{\mathrm{hi}} \subseteq \mathbb{R}^{n_{\mathrm{hi}}}\),
\(u \in U\subseteq \mathbb{R}^{m}\), and
\(f_{\mathrm{hi}} : [0,T] \times X_{\mathrm{hi}} \times U \to \mathbb{R}^{n_{\mathrm{hi}}}\).
Because high-fidelity system models are generally too expensive for reachable-set computation,
we introduce a lower-dimensional model on a shared state space, with state \(z \in Z \subseteq X_{\mathrm{hi}}\), and trajectory parameter \(k\in K\subseteq\mathbb{R}^{n_k}\), of the form
\begin{equation}
    \label{eq:lo_fid}
    \dot{z}(t) = f_{\mathrm{des}}(t,z(t),k).
\end{equation}
This model generates a parameterized family of candidate trajectories, and for each \(k\) the high-fidelity system is assumed to track the corresponding desired trajectory through an associated feedback controller~\citep{SK-SV-FB-MJR-RV:2020}. Note that neither~\eqref{eq:high_fid} nor~\eqref{eq:lo_fid} can include \textit{a priori} unknown disturbances for \emph{offline} FRS computation.

Because the planning model and the true closed-loop dynamics are not identical, RTD accounts for their mismatch through a tracking-error bound defined over the shared states~\citep{SV-US-SK-MJR-RV:2019}.
We introduce a tracking-error function \(g(t,k)\),
together with a disturbance-like signal \(d(\cdot)\in L_d\), where \(L_d = L^1([0,T],[-1,1]^{n_Z})\), and define the trajectory-tracking model componentwise as
\begin{equation*}
    \dot{z}_i(t)=f_{\mathrm{des},i}(t,z(t),k)+g_i(t,k)\,d_i(t),
\end{equation*}
for \(i=1,\dots,n_Z\). It is established in \citep[Lemma 12]{SK-SV-FB-MJR-RV:2020} that for any high-fidelity trajectory tracking any \(k\in K\), there exists \(d\in L_d\) such that the shared states satisfy the trajectory-tracking model.

The FRS is then defined as an over-approximation of the set of all shared states and associated trajectory parameters reachable by the trajectory-tracking model over the planning horizon~\citep{SK-SV-FB-MJR-RV:2020}:
\begin{equation}
\begin{aligned}
X_{\mathrm{FRS}} = \{(&\hat z,\hat k) \in Z \times K \mid\;
\exists\, t \in [0,T],\; z(0) \in Z_0,\; d \in L_d \\
&\text{s.t. }\
\dot z_i(\tau) = f_{\mathrm{des},i}(\tau,z(\tau),\hat k)
   + g_i(\tau,\hat k)\, d_i(\tau), \\
&z(0) = z_0,\quad z(t) = \hat z
\}.
\end{aligned}
\end{equation}

One can then define the following projection~\citep{SK-SV-FB-MJR-RV:2020} from state space to parameter space
\[
\pi_K(X')=\{k\in K \mid \exists z\in X' \text{ such that } (z,k)\in X_{\mathrm{FRS}}\},
\]
so that for an obstacle set \(X_{\mathrm{obs}}\subseteq Z\), the safe parameter set is \[K_{\mathrm{safe}} = \pi_K(X_{\mathrm{obs}})^C\], where \((\cdot)^C\) denotes the set complement.

The main strength of RTD is therefore that most of the difficult safety computation is shifted offline, while the online layer only solves an optimization problem over trajectory parameters. Its main limitation is equally clear: the price for safety is paid at a premium offline, where conservative tracking error bounds can cause safe, efficient trajectories to be incorrectly classified as unsafe and removed from the planner's feasible set.

\subsection{Mixed Monotone Reachability}
\label{subsec:LR-MMR}

Mixed monotone reachability provides a computationally efficient way to over-approximate reachable sets of nonlinear dynamical systems.
A detailed treatment of mixed monotonicity and its role in reachability analysis is provided in~\citep{SC:2020}.

A key property of this approach is that reachable set computation reduces to integrating an augmented system whose state encodes the lower and upper bounds of the original dynamics. This yields efficient over-approximations in the form of hyper-rectangles that can be propagated forward in time using standard numerical methods. Although such a representation may be conservative, its computational simplicity makes it well suited for real-time applications. Moreover, as demonstrated in Section~\ref{sec:experiments} and illustrated in Figure~\ref{fig:gap_comparison}, it is significantly less conservative than the worst-case bounds used in traditional RTD frameworks.

To formalize this construction, consider the nonlinear system
\begin{equation}
\dot{x}=f(x,u,w),
\label{eq:system}
\end{equation}
where \(x\in\mathbb{R}^n\) is the state, \(u\in\mathbb{R}^m\) the input, and \(w\in\mathbb{R}^p\) a disturbance.

Let \(\preceq\) denote the componentwise order on \(\mathbb{R}^n\), i.e., \(x\preceq y\) if \(x_i\le y_i\) for all \(i\). Given \(x,y\in\mathbb{R}^n\) with \(x\preceq y\), define the interval
\(
[x,y]:=\{z\in\mathbb{R}^n \mid x\preceq z\preceq y\}.
\)
For \(a=(x,y)\in\mathbb{R}^{2n}\), let \([\![a]\!]:=[x,y]\). We use the southeast order on \(\mathbb{R}^{2n}\),
\(
(x,x')\preceq_{\mathrm{SE}}(y,y')
\iff
x\preceq y \ \text{and}\ y'\preceq x',
\)
which implies
\(
(x,x')\preceq_{\mathrm{SE}}(y,y')
\iff
[y,y']\subseteq[x,x'].
\)
For an interval \([x,y]\), denote its endpoints by \(\lfloor[x,y]\rfloor=x\) and \(\lceil[x,y]\rceil=y\). Finally, for \(x,y\in\mathbb{R}^n\), let \(x_{[i:y]}\) denote the vector obtained from \(x\) by replacing its \(i^\regtext{th}\) component with \(y_i\).

Let
\[
\mathcal{F}=[\underline{\mathcal{F}},\overline{\mathcal{F}}]
\]
be an inclusion function for \(f\). That is, for any \(x\in[\underline{x},\overline{x}]\), \(u\in[\underline{u},\overline{u}]\), and \(w\in[\underline{w}, \overline{w}]\),
\[
f(x,u,w)\in
\mathcal{F}([\underline{x},\overline{x}],
[\underline{u},\overline{u}],
[\underline{w},\overline{w}]).
\]
Given such an inclusion function, the mixed monotone embedding system propagates the lower and upper bounds of the state via
\begin{align*}
\dot{\underline{x}}_i &=
\underline{\mathcal{F}}([\underline{x},\overline{x}_{[i:\underline{x}]}],
[\underline{u},\overline{u}],
[\underline{w},\overline{w}])_i,
\\
\dot{\overline{x}}_i &=
\overline{\mathcal{F}}([\underline{x}_{[i:\overline{x}]},\overline{x}],
[\underline{u},\overline{u}],
[\underline{w},\overline{w}])_i.
\end{align*}

This resulting \(2n\)-dimensional system evolves the interval bounds of the original dynamics and is monotone with respect to the southeast order. Consequently, if the initial state, inputs, and disturbances are over-approximated by intervals, the interval produced by the embedding system encloses the true reachable set.
Let
\[
\Phi^{\mathcal{E}}(t;[\underline{x}_0,\overline{x}_0],
[\underline{u},\overline{u}],
[\underline{w},\overline{w}])
\]
denote the embedding system state at time \(t\). If \(x_0\in[\underline{x}_0,\overline{x}_0]\), \(u(t)\in[\underline{u}(t),\overline{u}(t)]\), and \(w(t)\in[\underline{w}(t),\overline{w}(t)]\), then the interval produced by the embedding system over-approximates the reachable set of \eqref{eq:system}:
\begin{equation}
\phi(T;x_0,u,w)
\subseteq
[\![\Phi^{\mathcal{E}}(T;
[\underline{x}_0,\overline{x}_0],
[\underline{u},\overline{u}],
[\underline{w},\overline{w}])]\!]
\label{eq:mm_fundamental}
\end{equation}
for all $T \geq 0$~\citep{SC:2020}.

For general nonlinear systems, inclusion functions and corresponding embedding systems always exist, although deriving them analytically can be cumbersome. In this work we use the~\texttt{immrax} toolbox~\citep{immrax_paper}, which automates
the construction of inclusion functions and embedding systems
and enables efficient MMR computation in real time with just-in-time (JIT) compilation.

\section{Proposed Method: RTD-RAX}\label{sec:RTD-RAX}
This section describes the proposed \texttt{RTD-RAX} architecture. The central idea is to separate fast trajectory generation from execution-time safety certification. The RTD layer is retained as a rapid candidate generator, using a non-inflated offline FRS to avoid conservative planning, while mixed-monotone reachability is used online to certify the selected trajectory under the true system's tracking dynamics and real-time disturbances. If the candidate cannot be certified, the runtime layer either repairs it or blocks its execution.

\subsection{RTD Candidate-Generation Layer}
\label{subsec: RTDRAX-RTD}
The planning layer operates on states \(Z\) and trajectory parameters \(K\) as defined above. Each \(k\in K\) selects a desired trajectory and its associated control inputs.
Each candidate spans a horizon \(T=t_{\mathrm{plan}}+t_{\mathrm{stop}}\), consisting of a \emph{cruise phase} followed by a \emph{fail-safe phase}, ensuring that if replanning fails, the system can continue along a pre-determined safe terminating path.

The offline FRS encodes the set of states reachable while tracking these trajectories. Two variants are relevant: a \emph{standard} FRS, which includes tracking-error inflation, and a \emph{non-inflated} FRS, which removes that inflation.
At runtime, sensed obstacles are
converted into constraints \(q_i(k)\) against which we optimize over the set of possible trajectory parameters.
The planner then solves
\begin{equation}
    \min_{k\in K_{\mathrm{adm}}} J(k)
    \quad
    \text{s.t.}
    \quad
    q_i(k)\leq 0,\;\; i=1,\dots,N_{\mathrm{obs}},
    \label{eq:rtd_online_opt}
\end{equation}
where \(J(k)\) is the RTD objective (\emph{i.e.,} go toward a goal location), \(K_{\mathrm{adm}}\subseteq K\) is the admissible
parameter region, and \(N_{\mathrm{obs}}\) is the number of discretized obstacle points.

In standard RTD, the inflated FRS itself alongside the obstacle constraints serve as the execution-time safety certificate. In \texttt{RTD-RAX}, this role is reassigned: the non-inflated FRS is used to reject obviously unsafe parameters and generate a high-quality candidate quickly, while final execution is determined by an online reachable-tube computation. Thus the offline FRS no longer needs to encode all execution-time uncertainty conservatively in advance, while allowing the framework to account for disturbances in the loop.

\begin{figure}[h]
    \centering
    \includegraphics[
        width=.75\textwidth,
        trim={.7cm .85cm .9cm .3cm},
    ]{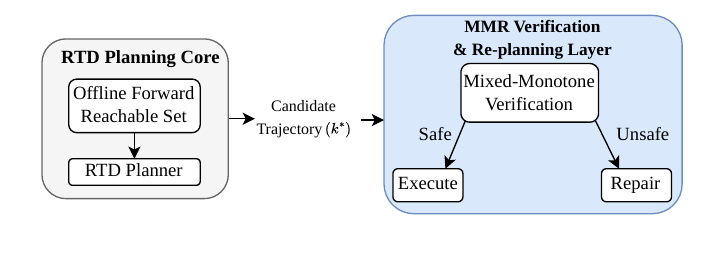}
    \caption{RTD-RAX architecture.}
    \label{fig:rtdrax_architecture}
\end{figure}

\subsection{Execution-Time Verification Layer}
\label{subsec:RTDRAX-RAX}

When \eqref{eq:rtd_online_opt} yields a candidate trajectory \(k^\star\), the architecture transitions from the RTD planning layer into the real-time safety verification layer. The purpose of this layer is to determine whether the specific trajectory proposed by the planner can be executed safely under the current uncertainty and disturbance assumptions.

The verification model takes the form
\begin{equation}
    \dot{x} = f(x,\,u_{k^\star}(t), w),
    \label{eq:rtdrax_verify_dyn}
\end{equation}
where \(x\in\mathbb{R}^n\) is the full system state, \(u_{k^\star}(t)\) is the control input prescribed by the candidate \(k^\star\) as a function of time, and \(w\in[\underline{w},\overline{w}]\subseteq\mathbb{R}^p\) represents bounded disturbance.

Given the verification dynamics \eqref{eq:rtdrax_verify_dyn} and the candidate parameter \(k^\star\), the runtime layer propagates a reachable tube using mixed-monotone reachability over the full execution horizon.

\subsubsection{Initial Interval Construction}

Safety verification begins from an uncertainty set
\(X_0 \subset \mathbb{R}^n\) centered at the current state estimate
\(\hat{x}_0\). This set captures the combined effect of state-estimation
uncertainty, the volume of the robot, and any additional safety margin required by the application.

We construct the uncertainty set as a hyper-rectangle
\begin{equation}
    x(0) \in [\hat{x}_0-\varepsilon,\;\hat{x}_0+\varepsilon],
    \label{eq:init_interval}
\end{equation}
where the vector \(\varepsilon\in\mathbb{R}^n_{\ge 0}\) specifies componentwise
bounds on the initial state uncertainty.
This initial interval is the mechanism through
which RTD-RAX incorporates execution-time uncertainty that would otherwise have
to be conservatively encoded into the offline FRS.

Disturbances acting during execution are modeled by a bounded set
\[
w(t) \in [\underline{w},\overline{w}],
\]
which may represent \textit{a priori} unkown environmental disturbances or otherwise unmodeled dynamics. Whereas traditional RTD frameworks have no mechanism through which to consider these into their notion of safe planning, by measuring these disturbances and incorporating them into the online reachable set computation, we can ensure safety in more general settings than traditional RTD.

\subsubsection{Embedding System Integration}
From the initial interval \eqref{eq:init_interval}, the verifier constructs an
embedding system
for the system dynamics~\eqref{eq:rtdrax_verify_dyn} to propagate the lower and upper bounds on each state component forward in time.

Let the embedding system state be
\[
(\underline{x}(t),\overline{x}(t)) \in \mathbb{R}^{2n}.
\]
At each time \(t\), this pair defines the interval
\[
X(t) = [\underline{x}(t),\overline{x}(t)]
\]
which over-approximates the set of states reachable by the true system for all
initial conditions \(x(0)\in X_0\) and all disturbance realizations
\(w(\cdot)\in[\underline{w},\overline{w}]\).
The embedding system is then integrated over the horizon using a fixed time step. At each sample time \(t_j\), the interval \[\mathcal{R}_j = [\underline{x}(t_j),\overline{x}(t_j)]\] bounds all states reachable at time \(t_j\). The ordered collection of such intervals over the horizon thus forms a reachable tube enclosing all trajectories.

\subsubsection{Workspace Projection and Collision Checking}
Let \(x_{\mathrm{pos}} \in \mathbb{R}^{n_p}\) denote the components of the state corresponding to the robot's position in the physical workspace (\emph{e.g.} planar position for ground vehicles or spatial position for aerial systems).
Projecting each reachable set \(\mathcal{R}_j\) onto these coordinates yields a sequence of axis-aligned bounding boxes
\[
\mathcal{B}_j =
[\underline{x}_{\mathrm{pos}}(t_j),\;
 \overline{x}_{\mathrm{pos}}(t_j)],
\]
which bound all possible robot positions at time \(t_j\).

The verifier then checks the bounding boxes as well as
swept interval hulls between each box
for intersection with every obstacle  and reports a collision at the earliest time at which any intersection is detected.
If no intersection is found, the candidate is certified safe over the entire horizon. If an intersection is found, the candidate is rejected before execution, and the earliest collision time is recorded for use by the repair procedure.

\subsection{Repair After Runtime Rejection}
\label{subsec: RTDRAX-REPAIR}

Because standard RTD formulations do not account for \emph{a priori} unknown runtime disturbances settings and safety is ensured through worst-case tracking error bounds, a notion of trajectory rejection is not needed. However, in under real-time disturbances, a trajectory returned by~\eqref{eq:rtd_online_opt} may be unsafe. In such cases, we implement a ``trajectory repair'' procedure that finds a similar but certifiably safe trajectory alternative in order to certify safety while still achieving the mission objective of the rejected trajectory.

When a candidate generated by the non-inflated RTD layer is declared unsafe, a sequence of corrective actions is applied until a safe trajectory is found. First, a \emph{speed backoff} reduces the forward velocity, thereby reducing the aggresiveness of the maneuver. If the resulting trajectory is still unsafe, a \emph{lateral push} adjusts the yaw rate parameter in opposition to the estimated disturbance. If neither correction yields a safe trajectory, \emph{constraint tightening} adds obstacle buffer to~\eqref{eq:rtd_online_opt} and re-solves. If a safe alternative is not found in time, the standard fail-safe maneuver is invoked.

\section{Experiments}\label{sec:experiments}
\begin{figure*}[t!]
    \centering

    \begin{subfigure}[c]{0.64\textwidth}
        \centering
        \includegraphics[
            width=\linewidth,
            trim={0.0cm 0.0cm 0.0cm 0.0cm},
            clip
        ]{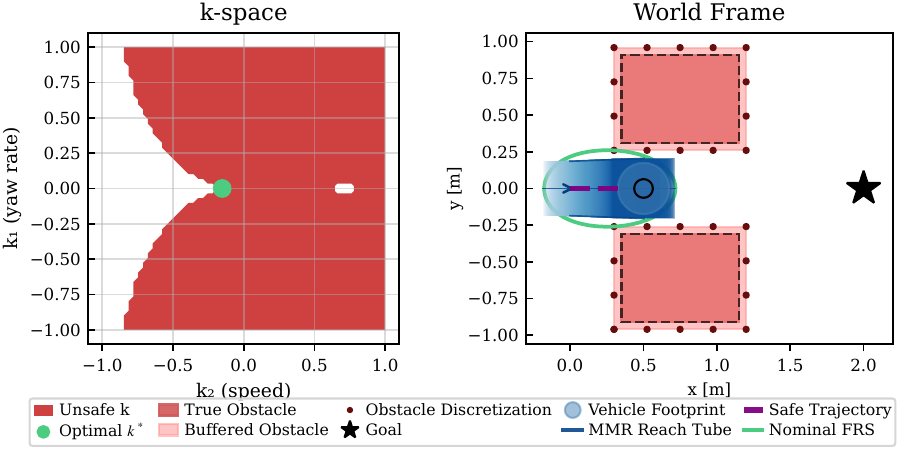}
        \caption{Narrow gap scenario. Left: feasible parameter space \(k\) (safe: white, unsafe: pink) with selected trajectory \(k^{*}\) in green. Right: certifiably safe trajectory through the gap.}
        \label{fig:case_gap}
    \end{subfigure}
    \hfill
    \begin{subfigure}[c]{0.31\textwidth}
        \centering
        \includegraphics[
            width=0.91\linewidth,
            trim={0.25cm 0.2cm 0.36cm 0.2cm},
            clip
        ]{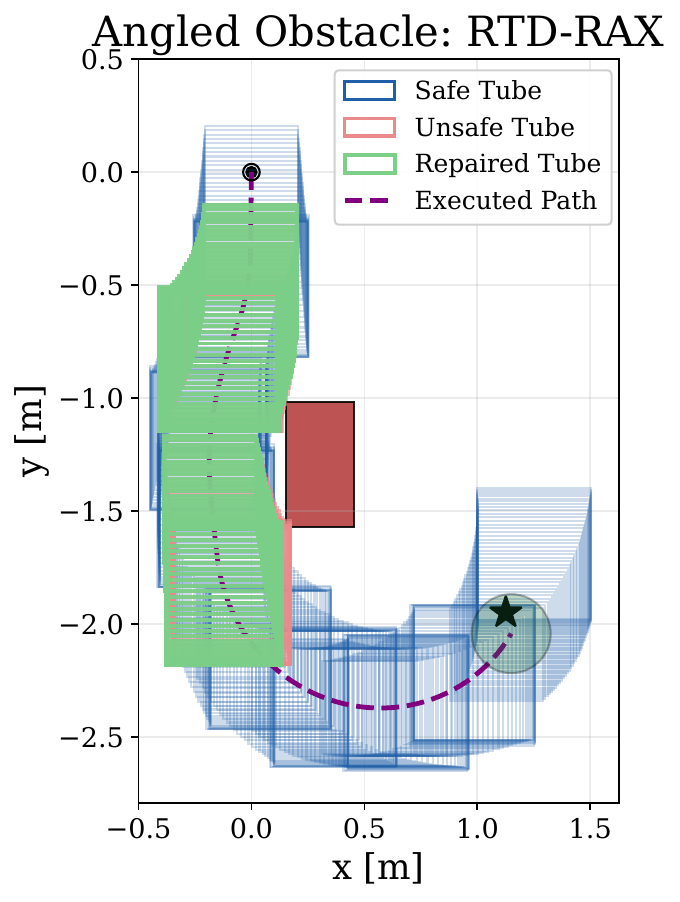}
        \caption{Angled-obstacle scenario with execution-time verification and repair. (See footnote~2 for additional details.)}
        \label{fig:case_angled_obstacle}
    \end{subfigure}

    \caption{Figures for the Narrow Gap scenario on the left and Angled Obstacle scenario on the right.}
    \label{fig:case_studies}
\end{figure*}
We demonstrate the utility of the proposed framework relative to traditional RTD approaches through three case studies.
To evaluate these scenarios, we consider a Turtlebot-scale ground robot modeled as a unicycle with state $(x, y, h, v)^\top \in \R^4$
governed by
\begin{equation}
    \dot{x} = v\cos h, \quad \dot{y} = v\sin h, \quad \dot{h} = \omega, \quad \dot{v} = a,
    \label{eq:unicycle}
\end{equation}
where the control inputs are the yaw rate $\omega$ and the longitudinal acceleration $a$.

Directly computing reachable sets for the full dynamics~\eqref{eq:unicycle}
over time leads to a high dimensional reachable-set problem that is computationally expensive for the offline SOS computation~\citep{SK-SV-FB-MJR-RV:2020}.
We instead utilize a Dubins' car as a planning model
defined on the shared state space \(Z \subseteq X_{\mathrm{hi}}\) with
\(
z = (x, y, h)^\top \in \R^3
\)
and dynamics
\begin{equation}
    \dot{x} = v_{\mathrm{des}}\cos h, \quad
    \dot{y} = v_{\mathrm{des}}\sin h, \quad
    \dot{h} = \omega_{\mathrm{des}} .
    \label{eq:planning_model}
\end{equation}

To further reduce the size of the control input space, the desired yaw rate and speed are not allowed to vary arbitrarily over time. Instead, they are held constant over each planning horizon and determined by a compact set of trajectory parameters
\(
k = (k_1, k_2)^\top \in [-1,1]^2
\) which map to the commanded inputs.
Because the parameters are fixed over each horizon, a single choice of $k$ uniquely determines a desired trajectory.

\subsection{Case Study 1: Narrow Gap}
We isolate the effect of offline FRS conservatism by constructing a scenario in which the standard RTD framework declares all trajectory parameters unsafe. Under \texttt{RTD-RAX}, the reduced conservatism of the non-inflated FRS allows feasible candidates to be considered while safety is enforced online.

The vehicle
encounters two symmetric rectangular obstacles with a
narrow
gap between them.
As a result, the standard RTD formulation declares the scenario infeasible. On the other hand, with \texttt{RTD-RAX}, the optimizer returns a feasible parameter \(k^\star\) corresponding to a straight-line trajectory through the gap. A mixed-monotone reachable tube is then propagated over the execution horizon, correctly certifying that this proposed path is safe.

This result highlights the division of roles in \texttt{RTD-RAX}: the offline FRS generates candidate trajectories, while execution-time verification certifies safety under the realized uncertainty. A maneuver rejected by standard RTD is recovered and certified safe.

Figure~\ref{fig:case_gap} illustrates this: the left panel shows a feasible region absent under the inflated FRS; the center panel shows that the same candidate would be classified as unsafe under standard RTD; and the right panel shows the resulting safe trajectory in the world frame.

\subsection{Case Study 2: Angled Obstacle with Runtime Repair}

We evaluate runtime verification and repair in a receding-horizon setting. The robot starts at the origin with heading $h_0 = -\pi/2$ and is forced to navigate around an obstacle to its goal location.
At each step, RTD generates a candidate trajectory using the non-inflated FRS, which is then certified or rejected and subsequently repaired by the runtime verifier.

The resulting trajectory is shown in Figure~\ref{fig:case_angled_obstacle}. When proposed trajectories are rejected by the verifier due to predicted collisions, the trajectory repair pipeline yields new trajectories that may be certified as safe.
This experiment highlights two properties of \texttt{RTD-RAX}. First, the runtime verifier identifies unsafe trajectories that satisfy offline FRS constraints, ensuring that safety is not compromised. Second, safe trajectory alternatives can be recovered rapidly through small adjustments. Moreover, we note that while standard RTD is capable of finding safe passage in this scenario, \texttt{RTD-RAX} finds shorter, more efficient trajectories.
\footnote{
Animations of this scenario alongside comparisons with standard RTD may be seen at:\newline
\url{https://evannsm.github.io/ws_RTD/case_studies}
}

\subsection{Case Study 3: Planning Against Disturbances}
Lastly, we consider the setting that standard RTD is least equipped to handle; that is, one in which significant disturbances alter our system dynamics. The vehicle traverses a \SI{6}{\meter} course with three offset gate obstacles. Before each gate, a disturbance patch pushes the vehicle toward the nearest obstacle.
This disturbance is not known \textit{a priori} to RTD but is assumed to be perfectly measured online by the verification layer.
\begin{figure*}[t!]
    \centering
        \centering
        \includegraphics[
        width=.98\textwidth,
        trim={0.2cm 0.2cm 0.2cm 0.2cm},
        clip
    ]{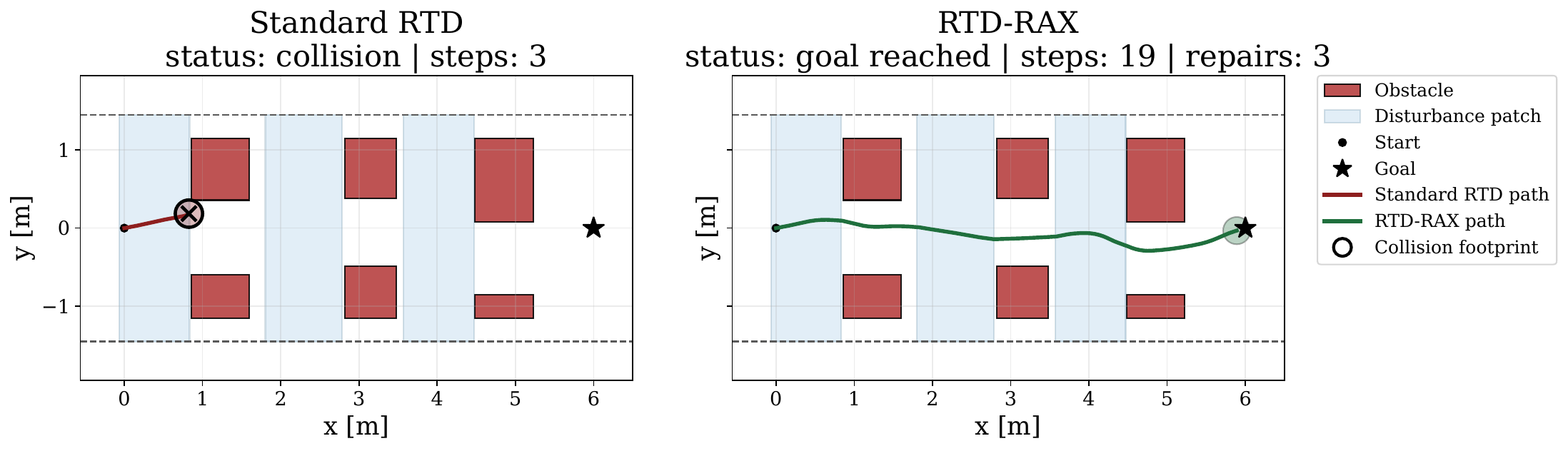}
    \caption{Disturbance-aware scenario.
    Left: standard RTD collides due to disturbances.
    Right: RTD-RAX with online certification accounts for disturbances and repairs unsafe candidates.}
    \label{fig:case_disturbance_compare}
\end{figure*}

As Figure~\ref{fig:case_disturbance_compare} demonstrates, in this scenario, the vehicle's footprint collides with the first obstacle during the third planning cycle. In contrast, \texttt{RTD-RAX} plans with the non-inflated FRS, computes disturbance bounds along each candidate, and certifies the resulting closed-loop motion online. Unsafe candidates are rejected and repaired, and the robot safely reaches the goal after 19 iterations. Moreover, as Table~\ref{tab:case3_timing} reports, standard RTD averages \(9.39 \pm 0.56\)\,\si{\milli\second} per planning iteration over 20 trials of this scenario, whereas \texttt{RTD-RAX} averages \(10.63 \pm 0.72\)\,\si{\milli\second}. This negligible difference of approximately \SI{1.24}{\milli\second} shows that runtime certification incurs minimal overhead, while allowing us to certify safety in uncertain environments, achieve more efficient trajectory tracking, and minimize needless failsave maneuvering.

\begin{table}[t]
    \centering
    \caption{Case Study 3: Mean and standard deviation per planning cycle over 20 trials.}
    \label{tab:case3_timing}
    \begin{tabular}{lcc}
        \toprule
        \textbf{Pipeline Step} & \textbf{Standard RTD (ms)} & \textbf{RTD-RAX (ms)} \\
        \midrule
        Constraint setup & $2.15 \pm 0.31$ & $2.41 \pm 0.22$ \\
        RTD solve & $6.48 \pm 0.38$ & $0.54 \pm 0.04$ \\
        Reference rollout & $0.73 \pm 0.17$ & $1.22 \pm 0.10$ \\
        \texttt{immrax} verify & $0$ & $2.23 \pm 0.14$ \\
        Repair loop & $0$ & $4.20 \pm 0.28$ \\
        \midrule
        \textbf{Total cycle} & $\mathbf{9.39 \pm 0.56}$ & $\mathbf{10.63 \pm 0.72}$ \\
        \bottomrule
    \end{tabular}
\end{table}

This experiment highlights the role of runtime certification under disturbances. The limitation we see now is no longer conservatism in the FRS, but the inability of the offline model to account for realized disturbances. \texttt{RTD-RAX} addresses this by incorporating measured disturbance bounds into an online reachable-tube computation, with only modest additional computational cost.

\section{Conclusion}\label{sec:conclusion}
This paper proposes~\texttt{RTD-RAX}, a safe trajectory design and runtime-assurance architecture for systems subject to \textit{a priori} unknown real-time disturbances.
The central idea is a complementary architecture: an offline FRS and optimization procedure rapidly generate candidate trajectories, while online certification accounts for real-time disturbances and model mismatch.
When a candidate cannot be certified as safe, a repair procedure seeks a nearby certifiably safe alternative before reverting to a fail-safe maneuver. The added verification and repair layers incur minimal computational overhead due to JIT compilation via~\texttt{immrax}.

We evaluate the framework in three case studies. In a narrow-gap scenario, standard RTD finds no feasible trajectory and triggers fail-safe behavior, whereas \texttt{RTD-RAX} correctly identifies a safe path. In an angled-obstacle setting, the runtime verifier identifies candidates that would become unsafe during execution and the repair loop recovers safe alternatives. In a disturbance-aware scenario, the verifier incorporates measured disturbances to enable safe execution where standard RTD leads to crashes.

Future work includes tighter coupling between the RTD optimization and the online reachable tube to enable gradient-based re-planning. Additional directions include richer disturbance and uncertainty descriptions, such as learned models or Gaussian processes,
as well as validation on hardware across a broad range of robotic platforms.

\bibliographystyle{plain}
\bibliography{2_arXiv_Version/ifacconf}

\end{document}